\documentclass[letterpaper, 10 pt, conference]{ieeeconf}  

\IEEEoverridecommandlockouts                              

\usepackage{amsmath} 
\usepackage{amssymb}  
\usepackage{graphicx}
\usepackage{float}
\usepackage{cite}
\usepackage{url}
\usepackage{booktabs}
\usepackage{xcolor}

\usepackage{algorithm}
\usepackage[noend]{algpseudocode}
\usepackage[hidelinks]{hyperref}
\usepackage{subfig}
\usepackage[font={small,it}]{caption}

\usepackage{amsthm}

\makeatletter
\let\OldStatex\Statex
\renewcommand{\Statex}[1][3]{%
  \setlength\@tempdima{\algorithmicindent}%
    \OldStatex\hskip\dimexpr#1\@tempdima\relax}
\makeatother

\urldef{\smith}\url{stephen.smith@uwaterloo.ca}
\urldef{\deiaco}\url{ryan.deiaco@uwaterloo.ca}
\urldef{\czarnecki}\url{kczarnec@gsd.uwaterloo.ca}

\newtheorem{theorem}{Theorem}

\title{\LARGE \bf
Learning a Lattice Planner Control Set for Autonomous Vehicles
}

\author{Ryan De Iaco, Stephen L. Smith, and Krzysztof Czarnecki
\thanks{This work was supported by the Natural Sciences and Engineering
Research Council.}
\thanks{The authors are with the Department of Electrical and Computer Engineering, University of Waterloo, Waterloo ON, N2L 3G1, Canada (\deiaco; \smith, \czarnecki)}}

\begin{document}

\maketitle
\thispagestyle{empty}
\pagestyle{empty}

\begin{abstract}

This paper introduces a method to compute a sparse 
lattice planner control set that is suited to a
particular task by learning from a representative dataset of vehicle paths.
To do this,
we use a scoring measure similar to the Fr\'echet distance and propose an
algorithm for evaluating a given control set according to the scoring measure.
Control actions are then selected from a dense control set according
to an objective function that rewards improvements in matching the dataset while
also encouraging sparsity. This method is evaluated across
several experiments involving real and synthetic datasets, and it is shown to
generate smaller control sets when compared to the previous state-of-the-art
lattice control set computation technique, with these smaller control sets maintaining
a high degree of manoeuvrability in the required task. 
This results in a planning time speedup of up to 4.31x
when using the learned control set over the state-of-the-art computed control
set. In addition, we show the learned control sets are better able to capture the driving
style of the dataset in terms of path curvature.

\end{abstract}

\section{Introduction}

\begin{figure}[thpb]
  \centering
  \includegraphics[scale=0.6]{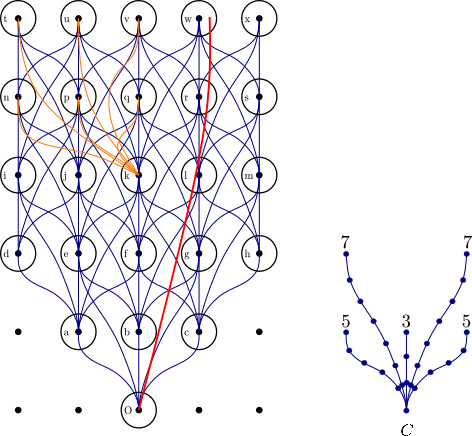}
  \caption{An example of a lattice graph, with labelled vertices. The control set is given
  by $C$, and each control action is labelled by the number of path points (excluding the origin point). An example control set from a different initial heading is given in orange at vertex $k$.
  The dataset path, $P_d$, is given in red.}
  \label{fig:dag_pic_a}
\end{figure}

A crucial portion of autonomous vehicle navigation is path planning. It is
important for autonomous vehicles to be able to quickly generate a collision-free,
kinematically feasible path towards their goal that minimizes the total cost of the path.
An algorithm commonly used in path planning is the lattice planner\cite{pivtoraiko_kelly_2005}.
The lattice planner is a graph-based approach to the path planning problem that reduces
the search space into a uniform discretization of vertices corresponding to positions and 
headings. Each vertex in the discretization is connected to other points by
kinematically feasible motion primitives, known as control 
actions\cite{pivtoraiko_nesnas_kelly_2009}. The lattice planner thus reduces
the path planning problem into a graph-search problem, which can be solved with A* or
any other appropriate graph search
algorithm\cite{pivtoraiko_knepper_kelly_2009, mcnaughton_urmson_dolan_lee_2011,
gu_2017, ziegler_stiller_2009}. An example of a lattice graph is shown in 
Figure~\ref{fig:dag_pic_a}.

\subsection{Contributions}

This work focuses on the task of leveraging data gathered from a particular task to optimize
a sparse set of motion primitives, known as a control set, by removing control actions that are
less important for planning paths similar to those in the dataset.
This sparse control set should be selected
such that it is specialized with respect to a given dataset; that is, it can reproduce a 
dataset of paths of an autonomous vehicle generated from human operation or demonstration.
The sparsity of the learned control set reduces the number of edges in the search graph, and thus 
allows for faster online path computation. In
addition, the learned control set should capture some characteristics of the driving style present
in the dataset, and the learned control set should not sacrifice path quality or manoeuvrability.
To learn such a control set, we require a way to measure how closely we can match paths from the
dataset using a lattice planner with the given control set, as well as a way to select a sparse subset.

In this work, the first contribution is a novel algorithm 
for finding the closest path in a lattice graph to a given path according to 
a modified version of the Fr\'echet distance. 
The second contribution is a method to select a sparse subset of a given control set that still
retains the ability to execute paths
in a given dataset, while also capturing the driving style present in the dataset. 
These algorithms are tested on both real human-driven data
as well as synthetic data, and compared to the state-of-the-art lattice control set
reduction technique\cite{pivtoraiko_kelly_2011}.

\subsection{Related Work}\label{sec:related_work}

In previous work, data-driven motion planning has often focused on learning
search heuristics or policies for the motion planner rather than learning the underlying
structure of the planner itself. Ichter et al. 
developed a method for learning a sampling distribution for RRT* motion planning\cite{ichter_harrison_pavone_2017}.
Imitation learning can also be used to learn a search heuristic 
based on previously planned optimal paths\cite{2017arXiv171106391C,
bhardwaj_choudhury_scherer_2017}.
Paden et al. have developed a method for optimizing search 
heuristics for a given kinodynamic planning problem\cite{paden_varricchio_frazzoli_2017}. 
Xu et al. used reinforcement learning to learn a control policy for quadcopters by training on MPC
outputs\cite{zhang_kahn_levine_abbeel_2016}.

For work involving lattice planner control set optimization,
Pivtoraiko et al. have developed a D*-like (DL) algorithm
for finding a subset of a lattice control set that spans the same reachability of the
original control set, but does so within a multiplicative factor of each original 
control action's arc length\cite{pivtoraiko_kelly_2011}. 
This algorithm does not rely on data, but instead relies on the structure
of the original control set to find redundancy. 
In contrast, our method attempts to leverage data for a particular application to
optimize the control set.
This paper uses the DL algorithm as the state-of-the-art comparison for the quality of the presented 
learning algorithm.

To optimize a planner, we require a measure of similarity
between two paths. This has been discussed in the field of path clustering
\cite{yuan_sun_zhao_li_wang_2016}, where measures such as the pointwise
Euclidean distance, Hausdorff distance\cite{chen_wang_liu_song_2011},
the Longest Common Sub-Sequence, and the Fr\'echet distance\cite{eiter_mannila_94}
are commonly used.

The work most closely related to the process of matching a specific path in a graph
is the map-matching problem\cite{wenk_2003, chen_driemel_guibas_nguyen_wenk_2011}. 
The problem entails finding a path
in a planar graph embedded in Euclidean space that best matches a given polygonal
curve according to the Fr\'echet distance. However, unlike our work, their algorithm
requires the full graph to be defined beforehand,
and cannot be used if the graph is implicitly defined in terms of the lattice control set.
Another similar problem is that of following a path in 
the workspace for a redundant manipulator\cite{oriolo_ottavi_vendittelli_2002, holladay_srinivasa_2016}.

In terms of driving style, Macadam gives
a broad overview of the driving task\cite{macadam_2003}. This paper focuses on the
properties of paths and not trajectories. For the driving style
of a given path, one of the most intuitive indicators is the vehicle's steering function,
which under the commonly used bicycle model\cite{polack_altche_dandrea-novel_fortelle_2017},
is directly related to path curvature. 
As such, curvature serves as a natural measure for comparing the
driving styles of different paths. A path with points of
high curvature corresponds to a a more aggressive steering function, and vice versa.

\section{Sparse Control Set Problem Formulation}

\subsection{Lattice Planner Preliminaries}

In this work, the robot navigates lattice points $(x, y, \theta)$ within
a subset $W \subset \text{SE}(2)$, 
discretized with $x$ and $y$ resolution $\Delta x$ and $\Delta y$, respectively, and 
with heading set $\Theta$\cite{pivtoraiko_knepper_kelly_2009}. Navigation between
lattice points in $W$ is done according to control actions present in a control set $C$.
For a given
control set $C$, each heading $\bar\theta \in \Theta$ has an associated control subset 
$C_{\bar\theta} \subseteq C$, and the control actions in that set can be applied
at any lattice point $(x, y, \bar\theta)$. This action results in a transition
to a point $(x', y', \theta')$ where the relative position $(x'-x, y'-y, \theta'-\bar\theta)$
is fixed for that action. Thus, the action connects all identically arranged pairs of
lattice points\cite{pivtoraiko_2012}. These connections define edges $E$, 
and these lattice points define vertices $V$ in a lattice graph $G$. 
An example of a lattice graph is given in Figure~\ref{fig:dag_pic_a}.
Each control action in $C$ has a corresponding path, and the path formed by the
concatenation of control actions in the lattice graph is denoted as $P_{l}$.

\subsection{Problem Formulation}\label{sec:problem_formulation}

Our main goal is to learn a sparse control set for a
lattice planner that retains the driving style that is present in a dataset.
We start with a dense control set and then incrementally generate a subset
by selecting the control actions that best improve the ability of the lattice planner to execute
the paths present in the dataset. In essence, we would like the dataset paths to become
approximate subpaths of lattice paths formed using our learned control set, as in 
Figure~\ref{fig:closest_path_example}. 
While optimizing in this way, however, we also want to encourage sparsity,
since larger control sets result in longer planning times. We can formally state the 
high-level problem.

\textbf{High-Level Problem.} Given a dense set of control actions $C$, and a dataset of
representative paths $D$, compute a minimal subset $\hat{C} \subset C$ that allows a lattice
planner to execute the paths present in $D$.

We split the high-level problem into two sub-problems. The first is measuring how well 
control sets match the dataset, and the second is optimizing the control set accordingly. 

\textbf{Subproblem 1.} Given a path $P_{d}$ and a set of control actions $\hat{C}$,
compute how well $\hat{C}$ executes $P_{d}$ according to a scoring measure $d$.

\textbf{Subproblem 2.} Given a scoring measure $d$, a dataset of paths $D$, and a dense
set of control actions $C$, select as small a subset of $C$, $\hat{C}$, as
possible that best executes $D$ in aggregate according to a scoring measure $d$.

Subproblem 1 is discussed in Sections~\ref{sec:distance_measure} and~\ref{sec:closest_path},
and Subproblem 2 is discussed in Sections~\ref{sec:control_set_optimization} and \ref{sec:clustering}.

\section{Sparse Control Set Generation}

\subsection{Scoring Measure}\label{sec:distance_measure}

To find the closest path generated by a lattice planner, $P_l$, to a path in the
dataset, $P_d$, we first need a scoring measure $d$ to evaluate the similarity of two paths.
For two paths parameterized by $t \in \left[0,1\right]$, and two monotonic increasing onto functions $\alpha, 
\beta : \left[0,1\right] \to \left[0,1\right]$, the Fr\'echet distance is given by
$$d_{f}(P_{d}, P_{l}) = \inf_{\alpha, \beta} \max\limits_{t \in [0, 1]}
  ||P_{d}(\alpha(t)) - P_{l}(\beta(t))||.$$
However, we would like a scoring measure that rewards $P_{l}$ for matching
$P_{d}$ closely at each point along the path, where points of comparison are
at equal arc lengths along each path. This means that rather than allowing any monotonic
increasing traversal of the paths during distance computation as in the Fr\'echet distance, 
the paths should be traversed at the same rate. In other words, if both paths were
traversed at a constant velocity, then the scoring measure should
compare points that are reached at the same time. When traversing both paths at the same 
rate, path pairs with a low score are likely to have similar driving styles
along the entire path. 

We therefore modify the Fr\'echet distance as follows.
For a given path to match $P_{d}$ with arc length $T$, a matching path $P_{l}$ that is at 
least as long as $P_{d}$, and where $t$ is an arc length parameterization of both paths, then 
our scoring measure, denoted as $d$, is
\begin{equation}
\label{dl}
d(P_{d}, P_{l}) = \max\limits_{t \in [0, T]} ||P_{d}(t) - P_{l}(t)||.
\end{equation}

\begin{figure}[thpb]
  \centering
  \includegraphics[scale=0.45]{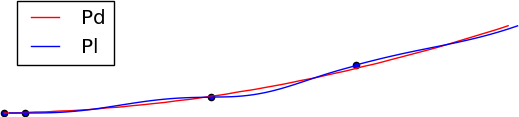}
  \caption{An example of the closest path found (blue) by Algorithm~\ref{alg:closestpath}
           with the red path as input.}
  \label{fig:closest_path_example}
\end{figure}

An advantage of using this measure instead of the Fr\'echet distance is that its simplicity 
allows for faster computation than the discrete Fr\'echet distance in a 
graph\cite{wenk_salas_pfoser_2006}.
Note that this scoring measure is no longer a distance metric, as it is
asymmetrical.
The fact that we perform a comparison only along the arc length of $P_d$ (and no further) 
is motivated as follows: rather than forcing the lattice path $P_l$ to be the same length
as $P_d$, we can plan $P_l$ to be arbitrarily longer and then truncate to the arc length
of $P_d$. This opens up a greater number of terminal lattice vertices when computing 
$P_l$, which we have found results in closer matching paths and faster runtime. 
The generation of $P_l$ is discussed in further detail in Section~\ref{sec:closest_path}.

Let us now assume that we are calculating $d$ for two discrete paths, sampled
with respect to arc length with segments of equal length $\delta$.
In Appendix~\ref{sec:practicalconsiderations}, we include implementation details, including
how to handle paths with length not integer-divisible by $\delta$.
Let $P_{d}$ contain $K$ sampled path points, $\{0, ..., K-1\}$, where the $0^{th}$
point is the origin. Let $P_{d}(k),
P_{l}(k)$ denote the $k^{th}$ path point of each respective path.
Then Equation~\eqref{dl} simplifies to
\begin{equation}
\label{dl_discrete}
d(P_{d}, P_{l}) = \max\limits_{k \in \{0, ..., K-1\}}
  ||P_{d}(k) - P_{l}(k)||.
\end{equation}
Equation~\eqref{dl_discrete} can be evaluated in $O(K)$ time.

Finally, for the algorithm discussed in the section below, we will need
to calculate $d$ between a control action $c \in C$
and a sub-path of an input path, where the sub-path starts
at path point $k_{1}$ and ends at path point $k_{2}$ of $P_d$.
In this case, both $c$ and the sub-path have $k_{2} - k_{1}$ segments between path points.
This is denoted by
\begin{equation}
\label{d_subpath}
d(P_{d}, c, k_{1}, k_{2}) = \max\limits_{k \in \{k_{1}, ..., k_{2}\}}
  ||P_{d}(k) - c(k-k_{1})||.
\end{equation}

\subsection{Closest Path Algorithm}\label{sec:closest_path}

In lattice planning, one typically searches for the shortest
path in the lattice graph to some goal point or region, where the lattice graph is
constructed according to a particular control set. 
However, to address Subproblem 1 of Section~\ref{sec:problem_formulation}, 
we instead wish to compute the path $P_l$ in the lattice
graph with minimum distance $d$ to a given dataset path $P_d$.
We assume both paths start at the origin $O$.

We propose Algorithm~\ref{alg:closestpath} to solve this problem.
To explain it, we first describe the input of a given
problem instance. We then discuss how we generate a search graph, followed by the
searching process. Finally, we analyze our proposed algorithm.\\

\subsubsection{Algorithm Input}

Figure~\ref{fig:dag_pic_a} illustrates example input to our algorithm. Here
we have a dataset path $P_d$ overlaid on top of a lattice graph constructed from an
input control set $C$.
The labelled vertices correspond to particular positions and headings
in space. We show a single heading across all vertices for visual clarity, except at
vertex $k$, which contains a control set for an alternative initial heading in orange.
The edges correspond to the underlying paths of the control actions that join points in space
according to $C$. The set $C$ is illustrated adjacent to the lattice graph. 
The underlying paths of each control action
are uniformly sampled with arc length $\delta$, and the corresponding number of path
points along each control action's path (excluding the origin point) are given as labels.

Each path is represented by a sequence of discrete path points, and as a result,
our scoring measure requires that the $k^{th}$ point along $P_{l}$ be compared with the
$k^{th}$ point along $P_{d}$ during computation.
To handle this, when generating the search graph we augment the lattice vertex with the
number of discrete path points $k$ along the path used to reach said lattice vertex.

\subsubsection{Search Graph}
We now describe the construction of the search graph.
As shown in Figure~\ref{fig:dag_pic_a}, there are multiple ways to reach vertex
$l$ in the lattice graph, some of which have different numbers of path points used along the
way. If $P_d$ contains $K$ path points, our search graph contains up to $K$ copies of each
vertex in the lattice graph to compute $d$. Each copy is differentiated by the number of path
points required to reach it. 

\begin{figure}[thpb]
  \centering
  \includegraphics[scale=0.6]{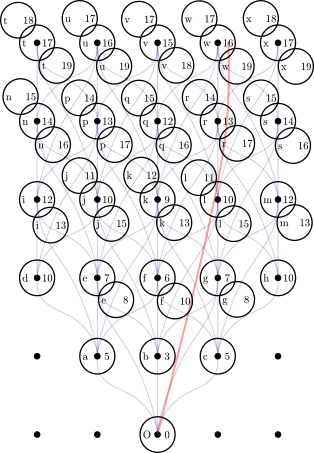}
  \caption{The search graph derived from Figure~\ref{fig:dag_pic_a}. Overlapping
  vertices correspond to the same point in space, but reached with a different number of path points. Some
  vertices are omitted for visual clarity.}
  \label{fig:dag_pic_b}
\end{figure}

These copies are illustrated in Figure~\ref{fig:dag_pic_b}. Revisiting vertex $l$, we can see that there 
are now three copies of $l$ in the 
search graph, each of which have a different value for the number of path points required
to reach it. The copies all correspond to the same point is space, but
with a different number of path points.

To illustrate why the search graph is useful, suppose we want to compute the $d$ scoring measure of
the control action from $(g, 7)$ to $(l, 10)$, as in Equation~\eqref{d_subpath}. This is shown in Figure~\ref{fig:dag_pic_c}.
The path points along this edge must be compared
to the path points 7 to 10 of $P_d$. This is shown by the dark green line segments between both
paths. The scoring measure of the control action from $(g, 7)$ to $(l, 10)$ is then the length of the longest
dark green line. However, if instead we wish to compute the $d$ scoring
measure of the control action from $(h, 10)$ to $(l, 15)$, we instead must compare to the path points 10 to 15
of $P_d$. This comparison is given by the light green lines between the paths.\\

\algrenewcommand\algorithmicindent{0.6em}%
\begin{algorithm}
\caption{\textsc{ClosestPath}($P_d, C, O, B$)}\label{alg:closestpath}
\begin{algorithmic}[1]
\footnotesize

\State $\text{bestEnd} \gets O$
\State $\text{costs, predecessors} \gets \text{HashTable()}$

\State $K = \text{length}(P_d)$
\State $V = \text{Array}(\text{HashTable}(), K)$
\State $V[0][O] = O$
\State $\text{costs}[O, 0] = 0$
\ForAll{$i \in {0, ..., K-1}$}
  \ForAll{$u \in V[i]$}
    \ForAll{$c \in C_{u.\theta}$}
      \State $(v, j) \gets \text{applyControlAction}(u, c, i)$
      \State $d_{u,v} \gets d(P_d, c, i, j)$

      \If{$d_{u,v} > B$}
        \State $\text{continue}$
      \EndIf

      \State $V[j][v] = v$
      \If{$\max(\text{costs}[u, i], d_{u,v}) < \text{costs}[v, j]$}
        \State $\text{predecessors}[v] \gets u$
        \State $\text{costs}[v, j] \gets \max(\text{costs}[u, i], d_{u,v})$
      \EndIf

      \If{$\text{costs}[v, j] < B \text{ and } j \geq K$}
        \State $\text{bestEnd} \gets v$
        \State $B \gets \text{costs}[v, j]$
      \EndIf

    \EndFor
  \EndFor
\EndFor

\State\Return $(\text{bestEnd}, \text{predecessors})$

\end{algorithmic}
\end{algorithm}

\subsubsection{Search Process}

Recall in Equation~\eqref{dl_discrete} that we are solving for the maximum pointwise distance between
$P_d$ and $P_l$. During our search, we seek to minimize this distance, i.e., find the 
closest path to $P_d$ in the search graph.
As we explore the search graph, we need to keep track of the
maximum pointwise distance computed along the closest path that reaches each search graph vertex. 

To solve this search problem, let us first denote the set of search graph vertices that require $k$
path points to reach them as $V_{k}$, and the collection of all $V_k$ as $V$ as shown in Line 4 of 
Algorithm~\ref{alg:closestpath}. All edges entering a vertex in $V_{k}$ 
come from some vertex in $V_{k'}$ such that $k'<k$. This then gives the vertices in the search graph a
topological ordering we can exploit, which we iterate through in Lines 7-20. Through each iteration,
successor vertices are found through applyControlAction(), which takes in a lattice vertex, a control action,
and the path point $i$ of that vertex, and outputs the successor lattice vertex as well as the 
resulting path point $j$ after applying the control action.
We can then apply a dynamic programming update for
each search graph vertex in every $V_{k}$ in increasing order of $k$ that computes the closest scoring measure
across all paths to each search graph vertex. If costs[] stores the best $d$ measure found so far
for each vertex, $U$ is the set of all predecessors of vertex $(v, j)$, and $d_{u,v}$ is computed
for the control action linking $(u, i)$ to $(v, j)$ according to Equation~\eqref{d_subpath}, then the update is given by
$$\text{costs}[v, j] = \min_{u \in U} \max(\text{costs}[u, i], d_{u,v}).$$
This update is shown in Lines 15 to 17.

\begin{figure}[thpb]
  \centering
  \subfloat[\label{fig:dag_pic_c}]{\includegraphics[scale=0.6]{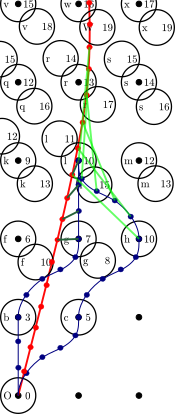}}\qquad
  \subfloat[\label{fig:dag_pic_d}]{\includegraphics[scale=0.6]{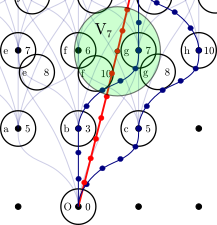}}\\
  \caption{(a) An example scoring measure computation to vertex $l$. 
  The light green line segments correspond to comparisons for the control action coming out of $(h, 10)$, and
  the dark green lines represent comparisons for the control action coming out of $(g, 7)$. (b) An illustration
  of a particular $V_k$ based on the greedy bound on the scoring measure. 
  }
\end{figure}

To reduce the number of vertices searched, we
compute an upper bound on the optimal $d$ scoring measure by greedily selecting control actions that minimize
the $d$ of the appropriate section of $P_l$. This bound, which we denote as $B$,
restricts the size of each $V_{k}$. This is illustrated in Figure~\ref{fig:dag_pic_d}. 
Only points within the shaded green circle can meet the scoring measure threshold $B$ given by the greedy
path. This means that $(g, 7)$ belongs to $V_7$, but $(e, 7)$ does not, as it is too distant. As a result,
outgoing control actions that reach $(e, 7)$ can be safely ignored, as any path that passes through
them is not as ``close" to $P_d$ as the greedily selected path. This is shown on Lines 12-13.
Recall the lattice resolution is given by $\Delta x$ and $\Delta y$. If we take $A = \Delta x \Delta y$,
the cardinality of each set $V_{k}$ is bounded by 
$\lceil\frac{B}{\Delta x}\rceil\lceil{\frac{B}{\Delta y}}\rceil|\Theta|
\in O(\frac{B^{2}}{A}|\Theta|)$.

Figure~\ref{fig:closest_path_example} gives an example solution using this method.
The algorithm takes in a path to follow, $P_d$, a control set, $C$, the origin of the lattice,
$O$, and the greedy bound, $B$, as input.
We start at the origin, iterate through each $V_k$ and apply the dynamic programming update described
above. The $V_k$ are populated during the graph search by successively applying control actions. 
The best scoring measure for each search graph vertex (as well as the associated predecessor 
vertex) is stored as the search progresses. 
This continues until all viable vertices have been searched, at which point we have found the
closest path to $P_d$ in the lattice graph.

\subsubsection{Algorithm Analysis}
We now analyze the correctness and runtime of Algorithm~\ref{alg:closestpath}.
In the algorithm, an empty entry in the costs hash table corresponds
to infinite cost.
To show the algorithm is correct, we show that when each vertex is processed in
topological order, the cost for said vertex is the minimum across all incoming paths.
We then discuss its runtime. Recall that $B$ is the greedy bound, $A=\Delta x \Delta y$,
$K$ is the number of points in $P_d$. In addition, we denote the maximum number
of path points across all control actions as $N$. The proof of the following result is
contained in Appendix~\ref{sec:proof}.

\begin{theorem}
Algorithm~\ref{alg:closestpath} is correct, and has runtime
$O(N\frac{B^{2}}{A}K|C|)$.
\end{theorem}

The runtime is heavily dependent on the quality of the bound $B$ provided, as a tight bound 
results in far fewer vertices to search. The $N$ factor is generally small relative to $K|C|$, 
so for a tight bound the runtime of the algorithm approaches $O(K|C|)$. This would be ideal,
as it corresponds to searching the control set at each point along the path.

\subsection{Control Set Optimization}\label{sec:control_set_optimization}

Now we present a method for optimizing the control set structure such
that it is best able to reproduce a given dataset. This is required to address Subproblem 2
in Section~\ref{sec:problem_formulation}. Recall that our objective is to select as
small of a subset as possible, $\hat{C}$, of an original dense control set $C$, while
still maintaining the ability to execute the paths in a given dataset. To accomplish
this our objective function should trade off between the sparsity of $\hat{C}$
and the ability of $\hat{C}$ to match the dataset. Recall that
the scoring measure in Equation~\eqref{dl_discrete} is denoted as $d$, the dataset
of paths as $D$, 
the initial dense control set as $C$, and the optimized control set as $\hat{C}$.
Define the set of all potential paths in the lattice as $\mathcal{P}(\hat{C})$,
and the parameter that trades off between sparsity and dataset matching as $\lambda$.
Then, our objective formulation is
\begin{equation}
\label{optimization_objective}
\min_{\hat{C} \subset C} \frac{1}{|D|} \sum_{P_{d} \in D} \min_{P_{l} 
  \in \mathcal{P}(\hat{C})} d(P_{d}, P_{l}) + \lambda\frac{|\hat{C}|}{|C|}.
\end{equation}

For each $P_d$, we are computing $d$
between $P_d$ and the closest path in the lattice graph constructed from $\hat{C}$, 
and summing over the entire dataset.
We normalize this value by the size of the dataset, to ensure consistency between
different dataset sizes. The second term penalizes the size of the learned control
set to encourage sparsity, and is normalized by the size of the initial dense control set. 
The $\lambda$ term is what trades off between sparsity
and dataset matching; a larger $\lambda$ results in a sparser control set, whereas a smaller
$\lambda$ allows the control set to fit the data more closely. In this sense, the $\lambda$
term acts as a regularizer in the objective function. Occam's Razor objective functions
that encourage simplicity are commonly used for tasks such as model selection or learning,
one of which is the Bayesian Information Criterion (BIC)\cite{murphy_2012}.

To perform the optimization, we start with a small control set $\hat{C}$. We then greedily
add the control action that results in the largest decrease in Equation~\eqref{optimization_objective},
and repeat until no control action can be added to further decrease the objective.
We use Algorithm~\ref{alg:closestpath} when computing the closest path according to $d$ as
required by Equation~\eqref{optimization_objective}.

\subsection{Clustering}\label{sec:clustering}

\begin{figure}[thpb]
  \centering
  \includegraphics[scale=0.4]{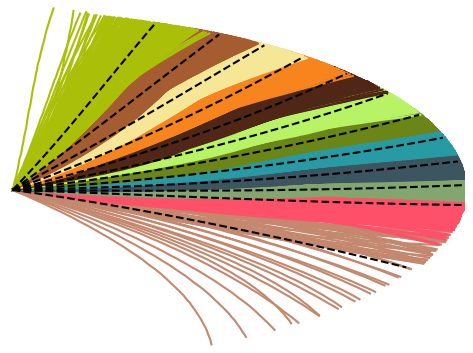}
  \caption{An example of the K-means clustering on a roundabout path dataset. Each cluster
  of paths has a different assigned colour, and the dotted line represents each cluster's mean
  path.}
  \label{fig:clustering_example}
\end{figure}

The optimization method above requires us to evaluate the objective
function for each available control action not yet within
$\hat{C}$ across all dataset paths to determine which control action is best to add. However, this
is computationally expensive. In addition, real world data often contains many similar paths. 
This is because there are often a limited number of ways 
to navigate a given scenario, and certain ways are more common than others. To alleviate
these issues, we first cluster the dataset using the K-means algorithm\cite{murphy_2012}.
To measure the distance between paths, we use the pointwise
Euclidean norm\cite{chen_wang_liu_song_2011}. An example of a clustering result is shown in 
Figure~\ref{fig:clustering_example}.

After clustering, we
bias our search process based on how well our learned control set is currently matching
each path cluster. Initially, each cluster has a large, equal weight.
Our optimization algorithm proceeds as follows:

\noindent\textbf{Control Set Optimization}
\begin{enumerate}
  \item Select a path cluster according to the selection weights, and randomly sample a subset of the path cluster and a subset of control actions.
  \item Compute the optimization objective for these subsets, adding each control action individually to 
  $\hat{C}$ and calling Algorithm~\ref{alg:closestpath} for each path in the cluster subset.
  \item Add the control action that decreases the objective the most to $\hat{C}$ permanently.
  Terminate if no control action improved the objective.
  \item Update the cluster selection weights with the resulting value of the optimization objective. Return to 1).
\end{enumerate}
This method focuses our optimization on clusters that are poorly matched.
Through this process, the optimization runs faster, and is more likely to match all types of
paths present in the dataset, rather than the most common ones.

\section{Results}

\begin{figure}[thpb]
  \centering

  \subfloat[\label{fig:dense_set}]{\includegraphics[scale=0.25]{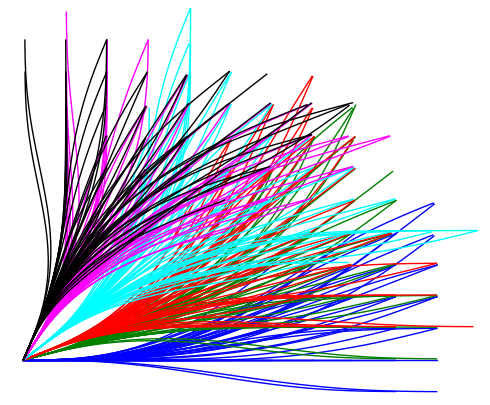}}\qquad
  \subfloat[\label{fig:pivtoraiko_set}]{\includegraphics[scale=0.25]{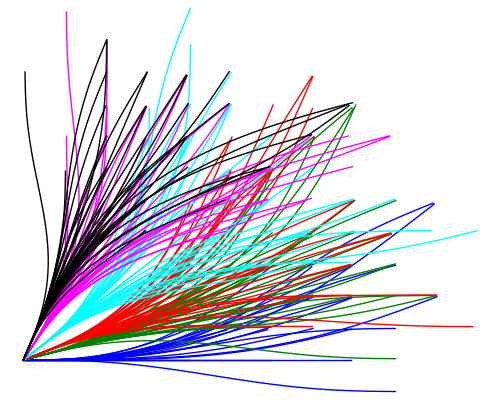}}\\
  \subfloat[\label{fig:l1_set}]{\includegraphics[scale=0.25]{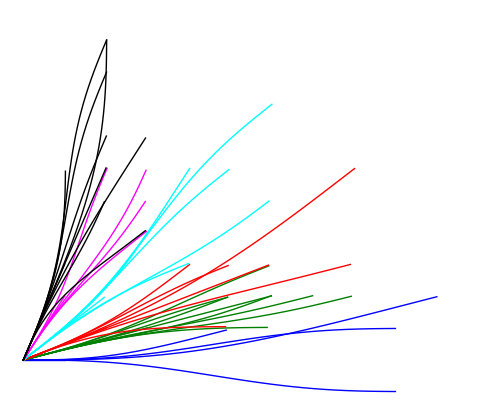}}\qquad%
  \subfloat[\label{fig:l2_set}]{\includegraphics[scale=0.25]{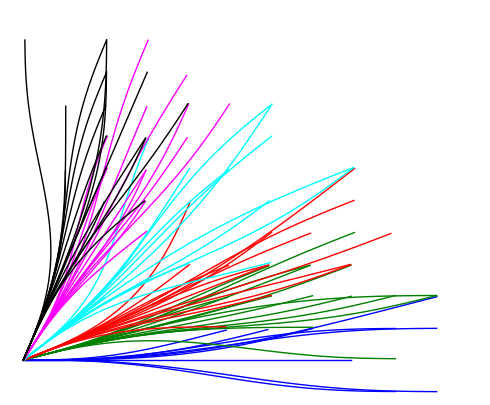}}%

  \caption{Comparison of the dense (a), DL\cite{pivtoraiko_kelly_2011} (b), $\lambda_{1}$ (c),
  $\lambda_{2}$ (d) control sets generated in Experiment 2.
  Each colour corresponds to a different $C_{\bar\theta}$.}
  \label{fig:control_sets}
\end{figure}

To evaluate our method, we devised three experiments. The first two used data
from human-driven trajectories around a roundabout, and the third used synthetic paths
created through randomly generated scenarios. In all three experiments, we performed an 85-15
split of the dataset between the training and test sets. The algorithms were written
in Julia. The source code for the
experiments can be seen at
\url{https://github.com/rdeiaco/learning_lattice_planner}.
For all experiments, the dense initial control set was a set of 
cubic spirals\cite{kelly_nagy_2003} arranged in a cone, generated for all $\theta \in \Theta$.
The endpoints of the control actions in the cone had a range of $x$ values between 0.4m and 4.0m,
a range of $y$ values between -2.0m to  2.0m, and $\theta$ values within [0, $\tan^{-1}(\frac{1}{3})$, 
$\tan^{-1}(\frac{1}{2})$, $\frac{\pi}{4}$, $\tan^{-1}(2)$, $\tan^{-1}(3)$]. 
These angles were chosen because they encourage straight line traversal between vertices in the
lattice graph, which improves path quality\cite{pivtoraiko_kelly_2005}.
The initial dense control set is shown in Figure~\ref{fig:dense_set}.

In each experiment we compared the performance of our learning algorithm to
the state-of-the-art lattice computation algorithm\cite{pivtoraiko_kelly_2011}.
The learning algorithm was run with $\lambda_{1}=0.311$ and $\lambda_{2}=0.0311$. 
These values were determined by logarithmically spaced grid search. Values of
$\lambda$ larger than this were found to generate control sets that were too sparse
with poor manoeuvrability. Swath-based collision checking was performed using a rectangular
vehicle footprint of length 4.5m and width 1.7m. Since the goal was not
necessarily reachable in the lattice graph, the lattice planner instead searched for
goal points that minimized the distance and heading difference from this goal.

\subsection{Experimental Setups}

\begin{figure}[thpb]
  \centering

  \subfloat[\label{fig:roundabout_view}]{\includegraphics[scale=0.2]{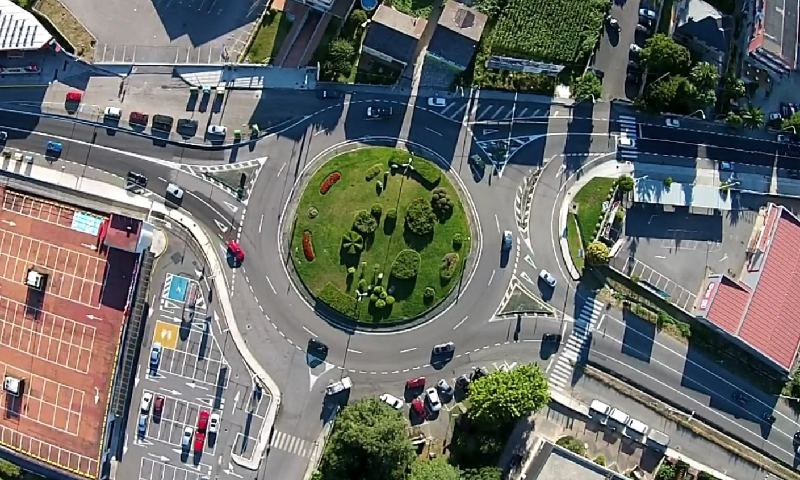}}\\
  \subfloat[\label{fig:generated_trajectories}]{\includegraphics[scale=0.32]{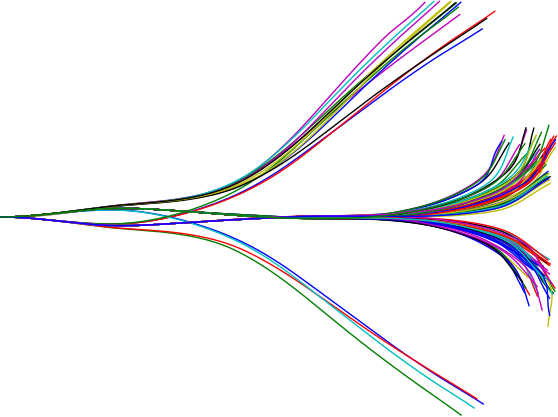}}\\

  \caption{(a) The roundabout the dataset was extracted from for Experiments 1 and 2.
  (b) The synthetic dataset generated using the Autonomoose planner.}
\end{figure}

\paragraph{Experiment 1: Roundabout Scenario}
The first experiment involved taking 213 paths in a roundabout dataset\footnotemark 
and sampling them at a constant arc length step size. 
The roundabout is illustrated in Figure~\ref{fig:roundabout_view}.
\footnotetext{Dataset obtained with permission from \href{www.datafromsky.com}{DataFromSky}.
The paths were extracted from cars driving through a European roundabout. The paths ranged
in length from 27.6 to 87.4m.}
The training portion of the dataset was then sliced
into 10m arc length slices using a sliding window with a 1m step size. These slices
were then taken as input to the clustering and optimization algorithms. 
This slicing method allows us to extract
as much information as possible from the dataset\cite{altche_fortelle_2017}.
To evaluate our learned control sets, we then took the test portion of our dataset
and constructed scenarios from each path. To do this, we took the test set path as
the lane centerline, with lateral offsets from the path forming the lane boundaries
in an occupancy grid. Finally, we used the endpoint of the test set path as the goal,
as well as the occupancy grid, 
and ran a lattice planner using each generated control set to compare the quality of each 
control set's planned paths. 

\paragraph{Experiment 2: Roundabout Lane Change Scenario}
The second experiment also involved the same training paths from the roundabout dataset,
except this time we added a second lane to the test set by extending the lateral offset
forming the lane boundaries. 
Rather than the goal being to travel to the end
of the original lane, the goal was changed to be the end of the adjacent lane. This
meant that the planner was required to perform a lane change, in order to demonstrate
that the learned control set could generalize to a situation not explicitly present in the
training set. The direction of the lane change was equally distributed between a left and
right lane change. Otherwise, scenario generation was the same as in Experiment 1.

\paragraph{Experiment 3: Synthetic Double Swerve Scenario}
For the third experiment, we generated 100 different lane structures by randomly sampling
clothoids of varying length and curvature connected to straightaways of varying length.
Next, a second lane was then added, along with an obstacle in the first lane.
The goal of this experiment was for the planner to perform a double swerve manoeuvre to avoid
the obstacle. We then
used the motion planner currently used on the University of Waterloo Autonomoose
self-driving car\cite{zhang_chen_waslander_yang_zhang_xiong_liu_2018} to generate the training set of 
synthetic paths. This dataset is shown in Figure~\ref{fig:generated_trajectories}.

\begin{table}[h]
\caption{Planning Runtime Results}
\label{table:results}
\begin{center}
\begin{tabular}{@{} l c c c c @{}}
\toprule
Experiment 1 & Dense & DL\cite{pivtoraiko_kelly_2011} & $\lambda_{1}$ & $\lambda_{2}$\\
\midrule
Control Set Size & 311 & 194 & 64 & 109 \\
Planning Speedup Ratio & 1.00 & 1.82 & 6.40 & 3.49 \\
Matching Differential (31 Scenarios) & - & -1 & +9 & +11 \\
\addlinespace
Experiment 2 & & & &\\
\cmidrule{1-1}
Control Set Size & 311 & 194 & 65 & 109 \\
Planning Speedup Ratio & 1.00 & 1.73 & 7.46 & 3.83 \\
Matching Differential (31 Scenarios) & - & +7 & +13 & +23 \\
\addlinespace
Experiment 3 & & & & \\
\cmidrule{1-1}
Control Set Size & 311 & 194 & 57 & 83 \\
Planning Speedup Ratio & 1.00 & 1.90 & 7.73 & 4.70 \\
Matching Differential (15 Scenarios) & - & +5 & +11 & +13 \\
\hline
\end{tabular}
\end{center}
\end{table}

\subsection{Experimental Results}

The results of all 3 experiments are shown in Table~\ref{table:results}. Here we can see
that the learned control sets are significantly smaller than both the dense control set
as well as the control set formed after performing the DL\cite{pivtoraiko_kelly_2011} 
lattice computation algorithm, illustrated in Figure~\ref{fig:control_sets}. 
Notably, this results in up to an approximately 7.5x planning speedup over
the dense set and up to a 4.31x planning speedup over the DL\cite{pivtoraiko_kelly_2011}
set when executing the test set. 

\begin{figure}[thpb]
  \centering
  \includegraphics[scale=0.45]{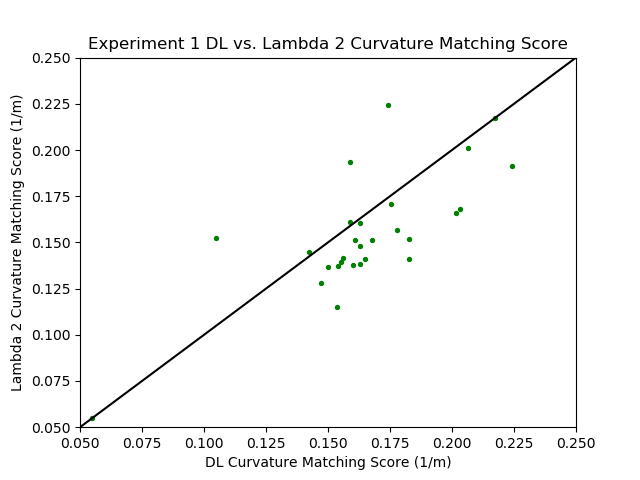}
  \caption{An example comparison of the curvature values between planned paths using
  the DL\cite{pivtoraiko_kelly_2011} and $\lambda_2$ control sets. 
  Each datapoint corresponds to a test scenario;
  below the straight line means that the $\lambda_2$ control set performed better.}
  \label{fig:curvature_comparison}
\end{figure}

To measure how well each control set matched
the dataset in terms of driving style, we computed the curvature at each point along
each planned path and dataset path as a proxy for the steering function, as discussed
in Section~\ref{sec:related_work}. Next, we computed the
maximum difference in curvature between each path point along the planned path and the dataset path. 
We call this the \emph{curvature matching score}. Afterwards, we compare
these curvature matching scores across the planned paths for each control set. The
value in the table reports the number of times a planned path had a lower maximum curvature
deviation than the dense set's planned path; a positive number denotes the control set was better
at matching more often than the dense set, and negative the opposite. A sample comparison
between the DL control set and the $\lambda_2$ control set is given in 
Figure~\ref{fig:curvature_comparison}. 

From this, we can see that the learned control sets
match the driving style (measured by curvature) of the dataset more closely than both the 
dense and DL\cite{pivtoraiko_kelly_2011} control sets, while also offering faster planning times. 
In addition, we can see
that as $\lambda$ gets smaller, the planned paths more closely match the data, at
the cost of a larger control set and slower planning times.

\begin{figure}[thpb]
  \centering
  \includegraphics[scale=0.5]{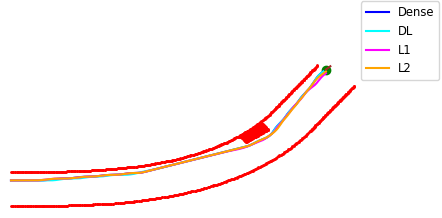}
  \caption{Comparison of the lattice planner paths for the dense,
  DL\cite{pivtoraiko_kelly_2011}, $\lambda_{1}$, and $\lambda_{2}$ control sets for
  one of the scenarios in Experiment 3.}
  \label{fig:gen_test_plots}
\end{figure}

Figure~\ref{fig:gen_test_plots} shows a sample planning run from
Experiment 3, comparing all 4 control sets. The red box denotes the obstacle for the
scenario.
We can see that all 4 planners were able to complete a plan to the goal state equally
well, which shows that the learned planners had no loss of manoeuvrability.

\section{Conclusions}

This work presents a novel method for learning a lattice planner
control set from a dataset of paths for a particular application. We
demonstrated its efficacy through experiments involving real and synthetic
data. The learned control sets are able to plan more quickly
than the state of the art control set generation technique, and
they better capture the driving style of the dataset
during the planning process.
In the future, we would like to explore combining learning the structure
of a lattice planner with learning the lattice planner's search heuristic, to see
if lattice planner performance can be improved even further for specific applications.
We would also like to extend this algorithm to handle trajectories rather than paths.

\appendix

\subsection{Proof of Theorem 1}\label{sec:proof}

We begin by proving correctness. To do this, we use induction on the vertices 
processed from $V$, as well as the fact that the vertices are processed 
in topological order.

Induction Assumption. For each vertex $u \in V_{k}$ processed from each $V_k \in V$,
we have that the cost assigned to $u$ is the minimal $d$ possible on
any path from the origin to $u$, when comparing said path to $u$ to the subpath $P_d(0:k).$ 

Base Case. The origin is the first processed vertex, and since $P_d$
starts at the origin, $d$ is zero, which is the correct distance.

Induction. Now, assume every processed vertex satisfies the induction assumption.
Suppose vertex $v$ is the current vertex to be processed. Since the algorithm processes
vertices in topological order, all potential predecessors of $v$ have already been processed,
and therefore satisfy the induction assumption.
By the dynamic programming update, taking $U$ to be the set of predecessors of $v$, we then
have that
$$\text{costs}[v] = \min_{u \in U} \max(\text{costs}[u], d_{u,v}).$$
Now, let $u'$ in $V_{0:k-1}$ denote the optimal predecessor of $v$. By the
update, we have that 
$$\text{costs}[v] \leq \max(\text{costs}[u'], d_{u',v}),$$
thus the induction assumption holds for $v$.

For runtime, Algorithm~\ref{alg:closestpath} iterates through a topological ordering of the search graph,
which can be thought of as $K$ groups of at most $\frac{B^{2}}{A}$ vertices. 
For each vertex in the topological ordering, we perform a dynamic programming update 
for each control action available to it. Across all headings, the total number of control 
actions available to any particular
vertex is $|{C}_{\bar\theta}|$, which in aggregate gives us $\sum_{\bar\theta \in \Theta}|C_{\bar\theta}| = |C|$. 
Each dynamic programming update calculates
$d$ for an edge, which takes $O(N)$ time. Combining, this
gives us a computational complexity of $O(N\frac{B^{2}}{A}K|C|)$.

\subsection{Practical Considerations}\label{sec:practicalconsiderations}

\emph{Arc Length Relaxation.}
Since the lattice control actions connect vertices in the lattice graph, a realistic 
application of this
method would require a small line segment length $\delta$, which would in turn
increase the size of $K$ required in each path matching calculation. 
To remedy this, we relax the requirement that each control action
has an arc length that is integer-divisible by $\delta$. 
This potentially results in a leftover portion of each control action that would be left
out of the closest path calculation. We overcome this by checking if the leftover portion of the
control action is greater than or equal to half of $\delta$. If it is, then we treat it
as a full line segment for $d$ computation. Otherwise, we ignore it. 
In practice, using a $\delta$ that is a $\frac{1}{4}$ of $\min(\Delta x, \Delta y)$ 
allows for good results.

\emph{Optimization Initialization.}
Finally, we initialize the learned control set with a single short, straight action for each
possible initial direction, to ensure that the closest path algorithm
can make forward progress when it encounters a point with any particular heading.

\addtolength{\textheight}{-10cm}   





\bibliography{citations}{}
\bibliographystyle{IEEEtran}

\end{document}